\documentclass[11pt,a4paper]{amsart}

\usepackage{amsmath}
\usepackage{graphicx}
\usepackage{times,epsfig} 

\usepackage{amssymb,amsthm}
\usepackage{color}
\usepackage{hyperref}
\usepackage{multirow}

\usepackage{etoolbox}
\makeatletter
\patchcmd{\@settitle}{\uppercasenonmath\@title}{}{}{}
\patchcmd{\@setauthors}{\MakeUppercase}{}{}{}
\makeatother

\begin{document}

\title{Unsupervised adaptation of brain machine interface decoders}

\author[G\"urel and Mehring]{Tayfun G\"urel$^{1,2}$ and Carsten Mehring$^{1,3}$\\
$^1$Bernstein Center Freiburg, Albert-Ludwig University of Freiburg, Germany\\
Current address: $^2$Imperial College London, Department of Bioengineering, London, UK\\
$^3$Imperial College London, Department of Bioengineering, London, UK\\
Tayfun.Guerel@bcf.uni-freiburg.de, mehring@imperial.ac.uk }

\begin{abstract}
The performance of neural decoders can degrade over time due to nonstationarities in the relationship between neuronal activity and behavior. In this case, brain-machine interfaces (BMI) require adaptation of their decoders to maintain high performance across time. One way to achieve this is by use of periodical calibration phases, during which the BMI system (or an external human demonstrator) instructs the user to perform certain movements or behaviors. This approach has two disadvantages: (i) calibration phases interrupt the autonomous operation of the BMI and (ii) between two calibration phases the BMI performance might not be stable but continuously decrease. A better alternative would be that the BMI decoder is able to continuously adapt in an unsupervised manner during autonomous BMI operation, i.e. without knowing the movement intentions of the user. 

In the present article, we present an efficient method for such unsupervised training of BMI systems for continuous movement control. The proposed method utilizes a cost function derived from neuronal recordings, which guides a learning algorithm to evaluate the decoding parameters. We verify the performance of our adaptive method by simulating a BMI user with an optimal feedback control model and its interaction with our adaptive BMI decoder. The simulation results show that the cost function and the algorithm yield fast and precise trajectories towards targets at random orientations on a $2$-dimensional computer screen. For initially unknown and nonstationary tuning parameters, our unsupervised method is still able to generate precise trajectories and to keep its performance stable in the long term. The algorithm can optionally work also with neuronal error signals instead or in conjunction with the proposed unsupervised adaptation.
\end{abstract}

\maketitle

%\footnote{Current address: Imperial College London, Department of Bioengineering, London, UK}
%\address{$^1$Bernstein Center Freiburg, Albert-Ludwig University of Freiburg, Germany\\}
%\address{$^2$Imperial College London, Department of Bioengineering, London, UK\\}

%\ead{\mailto{Tayfun.Guerel@bcf.uni-freiburg.de}, \mailto{mehring@imperial.ac.uk}}

\section{Introduction}
Brain Machine Interfaces (BMI) are systems that convey users brain signals into choices, text or movement \cite{Nicolelis2003,Lebedev2006,Donoghue2002,Wolpaw2002,Birbaumer1999}. Being still in development, BMI systems can potentially provide assistive technology to people with severe neurological disorders and spinal cord injuries, as their functioning does not depend on intact muscles. For motor control tasks, parameters of intended movements (e.g. movement direction or velocity) can be decoded from electrophysiological recordings of individual neurons \cite{Wessberg2000,Hochberg2006}, from local field potentials inside \cite{Mehring2003,Scherberger2005} and on the surface of the cerebral cortex \cite{Leuthardt2004, Mehring2004, Ball2009, Schalk2007, Pistohl2008} or from electrical fields on the scalp \cite{Wolpaw2004, Blankertz2003, Waldert2008}. The decoded parameters can be used for online control of external effectors \cite{Velliste2008, Ganguly2009, Schalk2008, Hochberg2006}.

The relation between recorded brain activity and movement is subject to change as a result of neuronal adaptation or due to changes in attention, motivation and vigilance of the user. Moreover, the neural activity-movement relationship might be affected by changes in the behavioral context or changes in the recording. All these nonstationarities can decrease the accuracy of movements decoded from the brain-activity. A solution to this problem is employing adaptive decoders, i.e. decoders that learn online from measured neuronal activity during the operation of a BMI system and that track the changing tuning parameters \cite{Taylor2002, Wolpaw2004}. 

Adaptive BMI decoders can be categorized according to which signals are employed for adaptation: \textbf{\emph{Supervised}} adaptive decoders use user's known movement intentions in conjunction with corresponding neuronal signals. During autonomous daily operation of the BMI systems, however, neither the user's precise movement intention nor his movement goal is known to the BMI decoder - otherwise one would not need a decoder. Therefore, supervised decoders can only adapt during calibration phases, where the BMI system guides the user to perform pre-specified movements. \textbf{\emph{Unsupervised}} adaptive decoders, in contrast, track tuning changes automatically without a calibration phase. They can for example benefit from multi-modal distributions of neuronal signals to perform probabilistic unsupervised clustering \cite{Blumberg2007, Vidaurre2011, Vidaurre2011b, Vidaurre2010}. Evidently much less information is available to the adaptation algorithm in the unsupervised case compared to the supervised case. Unsupervised decoders, hence, might not work for strong nonstationarities and might be less accurate and slower during adaptation. The third category, namely \textbf{\emph{error-signal based}} adaptive decoders, do not use an informative supervision signal such as instantaneous movement velocity or target position but employ neuronal evaluation (or error) signals, which the brain generates e.g. if the current movement of the external effector is different from the intended movement or if the movement goal is not reached \cite{Diedrichsen2005, Krigolson2008,Milekovic2012}. \emph{Unsupervised} and \emph{error-based} adaptive decoders are applicable during autonomous BMI control in contrast to \emph{supervised} adaptive decoders.

\subsection{Related work brain machine interfaces}
In earlier work, BMI research has already addressed online adaptivity issue. For instance, Taylor et al. \cite{Taylor2002} has proposed a BMI system, where individual neuron's directional tuning changes are tracked with online adaptive linear filters. Wolpaw and McFarland have shown that intended $2$-dimensional cursor movements can be estimated from EEG recordings \cite{Wolpaw2004}. In that study, they employed Least Mean Squares (LMS) algorithm to update the parameters of a linear filter after each trial. Later, Wolpaw et al. has also shown that a similar method can be used to decode $3$-dimensional movements from EEG recordings \cite{McFarland2010}. Vidaurre et al. have proposed adaptive versions of Linear Discriminant Analysis (LDA) and Quadratic Discriminant Analysis for cue-based discrete choice BMI-tasks \cite{Vidaurre2006, Vidaurre2007,Vidaurre2010}. These works employ supervised learning algorithms, i.e. they necessitate that the decoder knows the target of the movement or the choice in advance and adapts the decoding parameters. In other words, the employed methods know and make use of the \emph{true} label of the recorded neural activity.

More recently, DiGiovanna et al. \cite{DiGiovanna2009}, Sanchez et al. \cite{Sanchez2009} , Gage et al. \cite{Gage2005} have proposed co-adaptive BMIs, where both subjects (rats) and decoders adapt themselves in order to perform a defined task. This task is either a discrete choice task like pushing a lever \cite{Sanchez2009, DiGiovanna2009} or a continuous estimation task such as reproducing the frequency of the cue tone by neural activity \cite{Gage2005}. Gage et al. employ a supervised adaptive Kalman filter to update the decoder parameters that match the neural activity to cue tone frequency. DiGiovanna et al. and Sanchez et al. utilize a reward signal to train the decoder. The reward signal is an indicator of a successful completion of the discrete choice task. The decoder adaptation follows a reinforcement learning algorithm rather than a supervised one. Whether the target has been reached, however, in contrast to a fully autonomous BMI task, is known to the decoder.

Error related activity in neural recordings \cite{Falkenstein2000, Gehring1993} is very interesting from a BMI perspective. In both discrete choice tasks and cursor movement tasks, EEG activity has been shown to be modulated, when subjects notice their own errors in the given tasks \cite{Blankertz2003, Parra2003}. The modulation of the neural activity is correlated with the failure of the BMI task, and hence, can be used to modify the decoder model. With reliable detection of error related activity, the requirement for the decoder to know the target location could be removed. Instead, the error signal could be utilized as an inverse reward signal \cite{Rotermund2006, Mahmoudi2011}. An unsupervised, i.e. working in complete absence of a supervision or error signal, approach has also been taken for an EEG-based BMI binary choice task. Blumberg et al. have proposed an adaptive unsupervised LDA method, where distribution parameters for each class are updated by the Expectation-Maximization algorithm \cite{Blumberg2007}. More recently, unsupervised LDA has also been applied to an EEG based discrete choice task \cite{Vidaurre2011, Vidaurre2011b, Vidaurre2010}. Unsupervised LDA, however, is limited to finite number of targets. In other words, it can not be applied to BMI tasks where possible target locations are arbitrarily many and uniformly (or unimodal) distributed. Kalman filtering methods for unsupervised adaptation after an initial supervised calibration have also been proposed for trajectory decoding tasks \cite{Eden2004,Eden2004a,Wang2008}. These methods adapt by maintaining consistency between a model of movement kinematics and a neuronal encoding model.

\subsection{Optimal control theory for motor behavior}
Motor behavior and associated limb trajectories is most commonly and successfully explained by optimality principles that trade off precision, smoothness or speed against energy consumption \cite{Todorov2004}. This trade off is often expressed as a motor cost function. Within the optimality based theory motor behavior, open loop and feedback optimization compose two distinct classes of motor control models. The former involves the optimization of the movement prior to its start ignoring the online sensory feedback, whereas the latter incorporates a feedback mechanism and intervenes with the average movement when intervention is sufficiently cheap. Optimal feedback control (OFC) models explain optimal strategies better than open loop models under uncertainty \cite{Todorov2002}. OFC models also provide a framework, in which high movement goals can be discounted based on online sensory input flow \cite{Todorov2004}. Optimal feedback control usually accommodates a state estimator module, e.g. a Kalman filter, and a Linear-Quadratic controller, which expresses the motor command as a linear mapping of the estimated state \cite{Stengel1994}. The state estimator uses sensory feedback as well as the afferent copy of the motor command. The motor command is a feedback rule between the sensory motor system and the environment. OFC models obey the minimal intervention principle, i.e. they utilize more effort and cost for relatively unsuccessful movements in order to correct for the errors \cite{Todorov2002, Todorov2003}. Minimal intervention principle is also very important for the current work, as substantial deviations can result from both noise and a model mismatch between the organism and the environment. The non-minimal intervention, hence, can be interpreted as a sign of a possible model mismatch between a BMI user and the decoder. Recent evidence indicates that OFC should also model trial-by-trial and online adaptation in order to be plausible empirical evidence on motor adaptation \cite{Izawa2008, Braun2009}.

%\begin{figure}
%\begin{tabular}{cc}
%    \includegraphics[width=0.65\textwidth]{figure1.eps}  &
%      \includegraphics[width=0.35\textwidth]{figure2.eps}  \\
%\end{tabular}
%  \caption{Overview of the BMI user model and the environment model.}
% \label{fig:overw} 
%\end{figure}

\subsection{Scope and goals of our research}
During autonomous operation of a BMI system, the BMI decoder does not know the individual movement intentions of the subject nor the goal of the movement, apart from what can be derived from the measured brain activity and from sensing the environment. Hence, the decoder has no access to an explicit supervision signal for adaptation. We, therefore, developed an algorithmic framework for adaptive decoding without supervision in which the following adaptive decoding strategies could be implemented: 

\begin{enumerate}
\item Unsupervised, here the adaptation works using exclusively the neuronal signals controlling the BMI movements. 
\item Error signal based, the adaptation uses binary neuronal error signals which indicate the time points where the decoded movement deviates from the intended movement more than a certain amount.
\item Unsupervised + error signal based, the combination of the adaptive mechanisms of (i) and (ii).
\end{enumerate}

With a BMI system involving those strategies, lifelong changes in brain dynamics do not have to be tracked by supervised calibration phases, where users would go under attentive training. Instead, decoder adaptation would track possible model mismatches continually. The BMI user’s behavior could provide a hint to the decoder even without an explicit supervision signal. It is presumable that inaccurate movements result in corrective attempts, which in turn increase control signals and control signal variability. Optimal feedback control models, which widely explain human motor behavior, support this presumption as they would generate jerky and larger control signals under mismatches between the user’s and the system’s tuning parameters. Here, we develop a cost measure for online unsupervised decoder adaptation, which takes the amplitudes and the variations in the user's control signals into account (strategy i). Our unsupervised method incorporates a log-linear model that relates the decoding parameters to the cost via meta-parameters. Randomly selected chosen parameters are tested during also randomly chosen exploration episodes. In the rest of the time, the best decoding parameters according to the existing model (initially random) are used. The switch between these exploration and exploitation episodes is random and follows an $\epsilon$-greedy policy (see section~\ref{sec:methods}) . Harvested rewards for all episodes and associated decoding parameters compose the training data, from which meta-parameters are detected using the least squares method recursively. Note that we utilize the same algorithm for strategies (ii) and (iii). In strategy (ii), we employ the error signal as the cost instead of the derived one. In strategy (iii), a combination of both measures serves as the cost.

\section{Methods}
\label{sec:methods}
\subsection{Simulated task}

\begin{figure}
  \centering
    \includegraphics[width=0.58\textwidth]{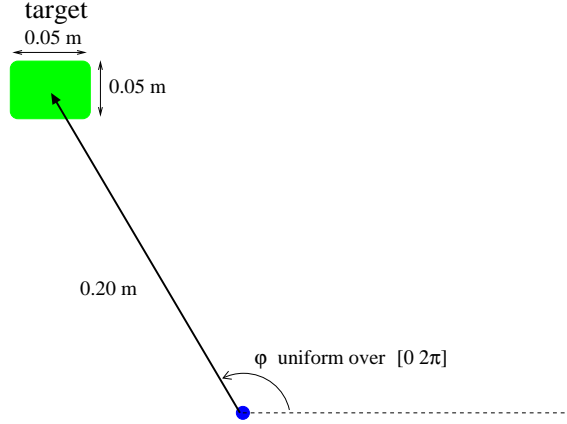}    
  \caption{BMI task: The user has to move the computer cursor towards the target. The target is $0.2$ meters away, at a random direction. The target location is decided by the user and unknown to the decoder, also for training purposes. The target has to be reached in $4$~seconds and the cursor has to stay on the target for at least $0.16$ seconds. Upon both success or failure, the user selects a new target.}
 \label{fig:task} 
\end{figure}

The user's task is to move a cursor on a $2$-dimensional screen from one target to the next. Each new target is located randomly on a circle of $0.2$~m radius around the previous target (figure~\ref{fig:task}). If the user reaches the target within $4$ seconds and stays there for $0.16$ seconds, the trial is successful. After an unsuccessful trial, the users selects a new random target. Upon success, the trial \emph{immediately} ends and the user selects a new target again. The state of the controlled system, i.e computer screen and cursor, at a discrete time step, $t$, is given by
\[
x_t = (p^1_t, p^2_t, v^1_t, v^2_t, g^1_t,g^2_t )^\mathrm{T}.
\]
where, $v^1_t,$, $p^1_t$ and $g^1_t$ are horizontal cursor velocity, cursor position and goal position, respectively. $v^2_t,$, $p^2_t$ and $g^2_t$ are the corresponding vertical state variables. The screen state evolves according to first order linear discrete time dynamics, 

\begin{equation}
x_{t+1} = A\,x_t \: + \:  B_d \,u_t,
\label{stateUpdate}
\end{equation}
where $u_t$ is the $C$-dimensional control signal and $B_d$ is a $6 \times C$ dimensional decoder matrix. We assume that the motor command, $u_t$,  affects only the cursor velocity directly. Therefore, $B_d$'s first $2$ and last $2$ rows are $0$:
\[
B_d= \left( \begin{array}{cccccc}
		0 &0& 0& 0& 0& 0\\
		0 &0& 0& 0& 0& 0\\
		b^{11}_d &b^{12}_d& & ...& & b^{1C}_d\\
		b^{21}_d &b^{22}_d& & ...& & b^{2C}_d\\
		0 &0& 0& 0& 0& 0\\
		0 &0& 0& 0& 0& 0\\
		\end{array}
	\right) .
\]
The state transition matrix $A$ models the temporal evolution of the screen state. It simply performs the operation $(p^1_{t+1}, p^2_{t+1} ) = (p^1_t, p^2_t )  + (v^1_t, v^2_t )$,
\[
A= \left( \begin{array}{cccccc}
		1 &0& 1& 0& 0& 0\\
		0 &1& 0& 1& 0& 0\\
		0 &0& 0& 0& 0& 0\\
		0 &0& 0& 0& 0& 0\\
		0 &0& 0& 0& 1& 0\\
		0 &0& 0& 0& 0& 1\\
		\end{array}
	\right) .
\]
Note that the goal position remains constant within a trial and it is left untouched by the linear dynamics of the screen state.
Including the goal position in the state vector, however, simplifies the formulation of control signal generation by the user model (section~\ref{usrModel}).

\begin{figure}
  \centering
    \includegraphics[width=0.98\textwidth]{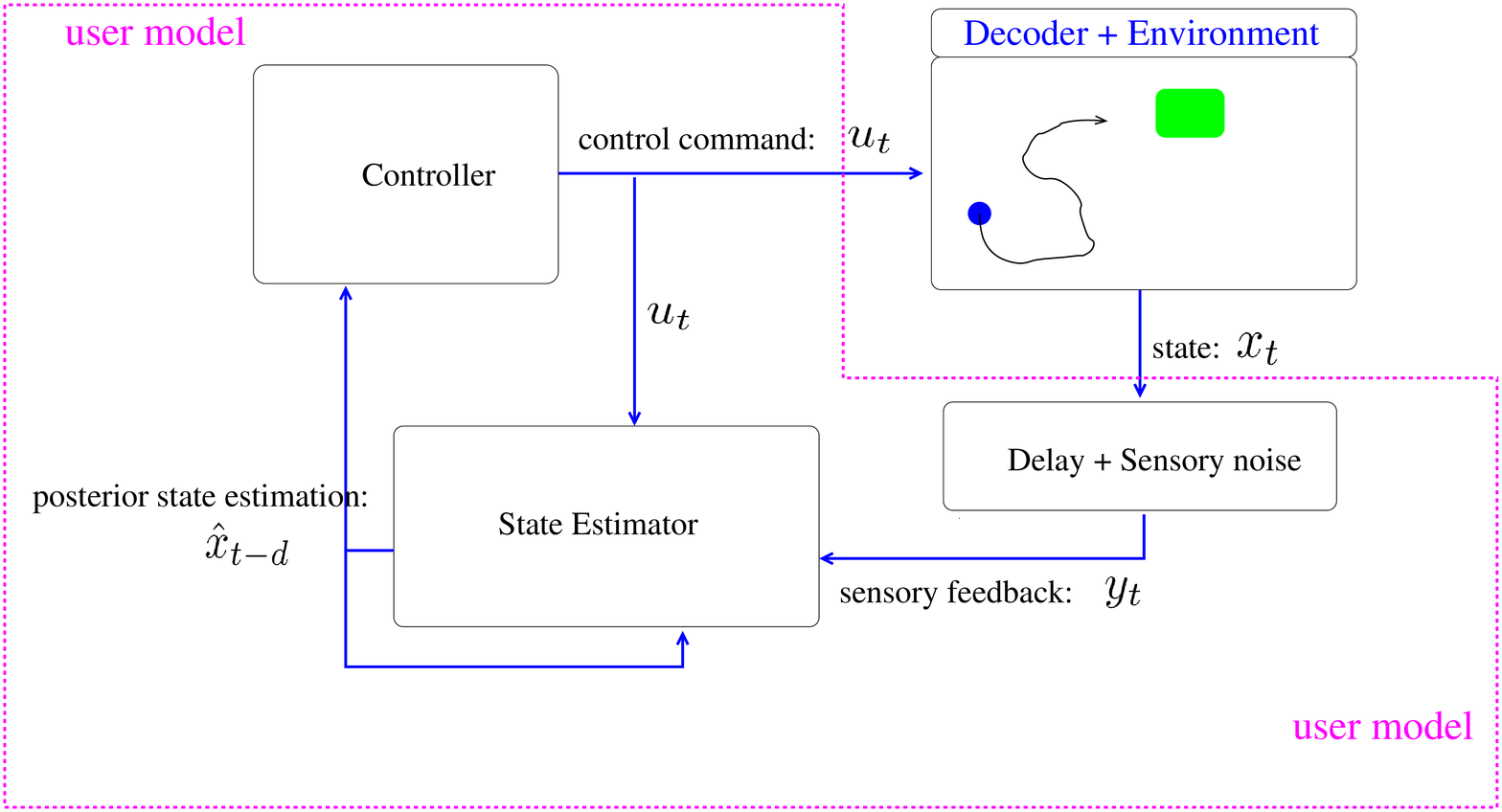}    
  \caption{ BMI user model and the environment. We model the BMI user with a stochastic optimal controller. The state estimator module is a Kalman filter that corrects the forward module estimation with sensory feedback. The controller generates the $C$-dimensional control command 
$u_t$, which is linearly converted to $2$-dimensional cursor movement by the decoder.}
 \label{fig:overw} 
\end{figure}

\subsection{User model: stochastic optimal controller}
\label{usrModel}
The BMI user is modeled as a stochastic optimal controller, who sends the $C$-dimensional control command $u_t$ at discrete time step $t$ (figure~\ref{fig:overw}) . The controller, i.e. the user, assumes that the screen state evolves according to a first order discrete time dynamics,

\begin{equation}
x_{t+1} = A\,x_t \: + \:  B_u \,u_t,
\label{stateUpdateUser}
\end{equation}
where $B_u$ is the user's estimation of the decoder matrix $B_d$,

\[
B_u= \left( \begin{array}{cccccc}
		0 &0& 0& 0& 0& 0\\
		0 &0& 0& 0& 0& 0\\
		b^{11}_u &b^{12}_u& & ...& & b^{1C}_u\\
		b^{21}_u &b^{22}_u& & ...& & b^{2C}_u\\
		0 &0& 0& 0& 0& 0\\
		0 &0& 0& 0& 0& 0\\
		\end{array}
	\right) .
\]

\noindent The BMI user perceives the state of the cursor with sensory delay and normally distributed zero-mean noise,

\[
y_t = H\,x_{t-d} \: + \: \eta_t, 
\]
where $d$ is the sensory delay in time steps and $\eta_t$ is the noise drawn from $\mathcal{N}(\mathbf{0},\Omega_\eta)$. The user observes a $4$-dimensional vector, $y_t$, which contains the velocity and position observations. $H$ is therefore:

\[
H= \left( \begin{array}{cccccc}
		1 &0& 0& 0& 0& 0\\
		0 &1& 0& 0& 0& 0\\
		0 &0& 1& 0& 0& 0\\
		0 &0& 0& 1& 0& 0\\
		\end{array}
	\right) .
\]
In computer simulations, we use a time step of $40$~ms. Sensory delay is set to $200$~ms, i.e. 5 time steps. We assume that all dimensions of $\eta_t$ are independently normally distributed with standard deviations of $(0.0004$~m, $0.0004$~m, $0.1$~m/s, $0.1$~m/s$)^\mathrm{T}$.

We model the control signals from the BMI-user as the output of a stochastic optimal controller. The BMI-user model aims at optimizing the cost function
\begin{equation}
\label{userCost}
J_u = \sum_t   \: (q \; \|g_t -p_t\|^2 \:+ \: r \; u^\mathrm{T}_t  u_t )\: ,
\end{equation}
$\|g_t -p_t\|$ stands for the euclidean distance between the $2$-dimensional cursor position and the goal position vectors. $q$ and $r$ are constants that account for the
relative weights of the two terms in the cost. The same cost expression can be written alternatively as,
\[
J_u = \sum_t   \: (x^\mathrm{T}_t \,Q_t\, x_t \:+ \:  u^\mathrm{T}_t \,R_t \,u_t )\: ,
\]
where $Q_t$ is a $6 \times 6$matrix that allows for the quadratic expression of the distance cost,
\[
Q_t= q\,\left( \begin{array}{rrrrrr}
		1 & 0&  0&  0& -1&  0\\
		0 & 1&  0&  0&  0& -1\\
		0 & 0&  0&  0&  0&  0\\
		0 & 0&  0&  0&  0&  0\\
	       -1 & 0&  0&  0&  1&  0\\
		0 &-1&  0&  0&  0&  1\\
		\end{array}
	\right),
\]
and $R_t = r \, I$. $Q_t$ and $R_t$ stay constant for all $t$ in our cost model, $Q_t = Q$ and $R_t = R$ for all $t$.

Assume that the stochastic optimal controller minimizes the cost by sending the optimal control command $u^*_t$ at every time step $t$. In fact, the optimal command is disturbed by noise. Here, we model the inherent noise in biological circuits with a $0$-mean normally distributed additive noise vector $\rho_t$,
\[
u_t = u^*_t + \rho_t.
\]
This noise consequently presents itself also as additive at state update in equation~\ref{stateUpdate}
\begin{eqnarray*}
x_{t+1} &= &A\,x_t \: + \:  B_d \,u_t\\
	      &= &A\,x_t \: + \:  B_d \,u^*_t + B_d \, \rho_t \\
	      	&= &A\,x_t \: + \:  B_d \,u^*_t + \omega_t, \\
\end{eqnarray*}
where $\omega_t \sim \mathcal{N}(\mathbf{0},\Omega_{\omega}) $. The problem of computing $u^*_t$ is known as Linear-Quadratic-Gaussian (LQG) control  and can be recursively solved by an interconnected Linear-Quadratic-Regulator \cite{Stengel1994, Todorov2005}, 

\begin{eqnarray*}
%\label{contrSignal}
u^*_t &=& - L_t\,\hat{x}_t \\
&& \nonumber  \\
%\label{Lcompute}
L_t	 &=& (R + B_u^\mathrm{T} S_{t+1} B_u)^{-1} B_u^\mathrm{T} S_{t+1} A \\
&& \nonumber  \\
S_t &=& Q_t + A^\mathrm{T} S_{t+1}(A - B_u \,L_t),\nonumber 
\end{eqnarray*}
and a state estimating Kalman Filter, 
\begin{eqnarray}
 \hat{x}_{t+1}& = & A\, \hat{x}_t + B_u \,u^*_t  \;+\;K_t \,(y_t -H \,\hat{x}_t) \label{KF1}\\
&& \nonumber\\
K_t &=& A \, \Sigma_t H^\mathrm{T} (H \Sigma_t H^\mathrm{T} + \Omega_\eta)^{-1} \label{KF2}\\
&& \nonumber \\
\Sigma_{t+1} & = & \Omega_{\text{FW}}  + A \, \Sigma_t \, A^\mathrm{T}  - K_t H \Sigma_t A^\mathrm{T}. \label{KF3}
\end{eqnarray}

\noindent Here, $\Sigma_{t}$ is the covariance estimate of the state vector variable $x_t$ and $\hat{x}_t$ is estimate of its mean value posterior to noisy observation $y_t$. $\Omega_{\text{FW}}$ is the covariance of the noise associated with the forward model prediction. Kalman filter above is a model for the state estimator in user's motor control circuitry. $B_u$ is the user's estimation for $B_d$. When $B_u$ deviates from $B_d$, the user's control signals are not optimal anymore. Above equations assume that the sensory delay equals to 1 time step. Larger sensory delays, e.g. $d$ time steps, can be realized by using an augmented state vector, $\tilde{x}_t$, which contains $d+1$ states together \cite{Todorov2002, Braun2009},
\[
\tilde{x}_t = (x^\mathrm{T}_t, x^\mathrm{T}_{t-1}, \dots, x^\mathrm{T}_{t-d})^\mathrm{T}.
\]
State transition and observation matrices are redefined for the augmented state space,
\[
\tilde{A} = \left( \begin{array}{rrrrr}
		A & 0&   \cdots &  &0\\
		I & 0&   \cdots & &0\\
		0 & I&   \cdots &0&0\\
		\vdots & \vdots& \ddots&\vdots&\vdots \\
		0 & 0&   \cdots &  I&0\\
		\end{array}
	\right), \text{   \:\:  }\tilde{H} = (0,\,\cdots,\, 0,\, H)\text{\:  and   \:}  \tilde{B_u} = \left( \begin{array}{l}
														B_u\\
														0\\
														\vdots\\
														0
														\end{array}
														\right),
\]
in order to satisfy the system dynamics, $\tilde{x}_{t+1} = \tilde{A}\, \tilde{x}_t + \tilde{B_u} \,u_t $. Kalman filter equations~\ref{KF1},~\ref{KF2} and~\ref{KF3} are also modified according to augmented states and system parameters: prior state and covariance estimations before the delayed observation, i.e. $y_{t+1}= H\,x_{t+1-d}+ \eta_{t+1}$, are computed using the subject's forward model,
\begin{eqnarray*}
\tilde{x}^+_{t+1}&=& \tilde{A}\, \hat{\tilde{x}}_t + \tilde{B_u} \,u^*_t \\
& &\nonumber\\
\tilde{\Sigma}^+_{t+1}  &=& \tilde{A}\, \hat{\tilde{\Sigma}}_{t}\,\tilde{A}^\mathrm{T} + \tilde{\Omega}_{\text{FW}},\\
\end{eqnarray*}
%where $\tilde{\Omega}_{\text{FW}}$ is the covariance of the noise associated with the forward model prediction. 
\noindent Posterior state and covariance estimates are similarly computed using the Kalman gain matrix $\tilde{K}_{t+1}$,
\begin{eqnarray*}
\tilde{K}_{t+1} &=& \tilde{\Sigma}^+_{t+1} \tilde{H}^\mathrm{T} (\tilde{H} \tilde{\Sigma}^+_t \tilde{H}^\mathrm{T} + \Omega_\eta)^{-1}\\
& & \nonumber\\
\hat{\tilde{x}}_{t+1}  &=&  \tilde{x}^+_{t+1} \;+\;\tilde{K}_{t+1}\,(y_{t+1} -\tilde{H} \,\tilde{x}^+_{t+1})\\
& &\nonumber\\
\hat{\tilde{\Sigma}}_{t+1}  &=& (I - \tilde{K}_{t+1}\ \tilde{H})\, \tilde{\Sigma}^+_{t+1}.
\end{eqnarray*}
\noindent Note that in our simulations, $\tilde{\Omega}_{\text{FW}}$ is set to a diagonal matrix, whose first $4$ diagonal entries are the squares of the noise standard deviations, $(0.0025$~m, $0.0025$~m, $0.625$~m/s, $0.625$~m/s$)^\mathrm{T}$, and the remaining entries are $0$.

\subsection{Decoder models}
The decoder is modeled by $B_d$, i.e. it decodes velocity information from the neuronal control signal $u_t$. This decoder matrix might deviate from the user's decoder matrix $B_u$, on the basis of which he generates his control signals. Therefore, the proposed adaptive decoders adapt their $B_d$ according to $B_u$. In the current section, we describe three decoders: Our recursive least squares (RLS) based learning algorithm with \emph{unsupervised} and \emph{error-signal based} cost functions as well as a \emph{supervised} RLS filter for performance comparison.

\subsubsection{Unsupervised learning algorithm}
For \emph{unsupervised} and \emph{error-signal based} decoder adaptation, we define a cost function and estimate $B_u$ by optimizing the proposed cost function. In the unsupervised setting, the cost is associated with control signal,

\label{sec:us}

\begin{equation}
\label{usCost}
J^d_n = \sum_{t=n-T+1}^n  u^\mathrm{T}_t  u_t,
\end{equation}
Here, $u^\mathrm{T}_t$ stands for the transpose of the control command vector. $t$ and $n$ are indices over time steps. $T$ is the number of time steps in the control signal history for computing the cost function. Note that the decoder needs to know only the control signal, $u_t$, in order to compute the above cost function. This cost function reflects the control-related term of the user's cost function (equation~\ref{userCost}). The value of the cost function is expected to be high, if the user aims at correcting the movement errors which result from a model mismatch between the user and the decoder, i.e. between $B_u$ and $B_d$.

We name the cost in equation~\ref{usCost} \emph{amplitude cost}, as it is based on the amplitudes of the control commands. We, however, propose a further cost function that can be utilized for decoder adaptation, namely \emph{deviation cost}. Deviation cost uses the variances of the control signals across time instead of the summed squared norms of the control commands,
\begin{equation}
\label{usCost2}
J^d_{dev} (n)=  \sum_{c=1}^C \: \: \sum_{t=n-T+1}^n  ({u^c_t} -\bar{u^c})^2 ,
\end{equation}
where $c$ is an index over control channels ${u^c_t}$ is the control command at channel $c$ at time step $t$. $\bar{u^c})$ is the mean value of 
${u^c_t}$ for channel $c$ across the interval $[n-T+1, \,n ]$. A weighted sum of the above costs can also be used as the cost function,
\begin{equation}
\label{usCost3}
J^d_{ampl+dev} (n)= \sum_{t=n-T+1}^n  u^\mathrm{T}_t  u_t + \: \: Z \: \: \sum_{c=1}^C \: \: \sum_{t=n-T+1}^n  ({u^c_t} -\bar{u^c})^2 ,
\end{equation}
where $Z$ is a constant for weighting the contributions from each individual cost type.

Alternatively, in case neuronal evaluation signals (i.e. error signals) are available in the recordings, we use the number of errors over a finite number of discrete time steps as cost,
\begin{equation}
\label{errorCost}
J^d_n = \sum_{t=n-T+1}^n err_t.
\end{equation}
We simulated the neuronal error signal by assuming that neuronal error signals are generated if the deviation between intended and performed velocities exceeds a certain amount,
\[
err_t = \begin{cases}   1, & \text{   for   } cos(v^*_t , v_t) \leq cos(20^{\circ} )\cr
			    			0,  & \text{   for   }  cos(v^*_t , v_t) > cos(20^{\circ} ),
	\end{cases}
\] 
where $v^*_t$ is the intended velocity. $err_t$ is swapped probabilistically with a probability of $\kappa$ in order to reflect the reliability of error signals.  Note that similar binary \emph{movement mismatch events} are also recorded in human ECoG \cite{Milekovic2012}, though $20^{\circ}$ in our simulation was arbitrarily choosen (see discussion).

We assume a log-linear model for the decoder cost. Let $\beta$ be the parameter vector generated by the horizontal concatenation of the third and fourth rows in $B_d$ matrix, i.e. $\beta = [B_{d3}, B_{d4}]$. The model estimates the decoder cost as,
\begin{equation}
\label{estimatedDecoderCost}
\hat{J}^d = \exp ( -[\beta^\mathrm{T} \; b]\, w),
\end{equation}
where $b$ is a constant bias value concatenated to the flattened decoder parameter $\beta$ and $w$ is the column vector of the meta-parameters of this log-linear model. We denote the $-\log$ of the decoder cost by $\ell$, 
\[
\hat{\ell_n} = -\log(\hat{J}^d_n) = [\beta_n^\mathrm{T} \; b]\,w_n =  {\beta_n^\prime}^\mathrm{T} w_n.
\]
Let $[\beta^\mathrm{T} b] = {\beta_n^\prime}^\mathrm{T}$. Here, the task is to learn $w$ from explored $\beta$ and $J^d$ collections and to simultaneously optimize $\beta$ for a given $w$. Note that for a given $w$, the cost-minimizing $\beta$ would go the infinity, since $-\log$-cost linearly depends on $w$. Therefore, the minimization is performed on the unit circle, i.e. $|\beta| =1$. The motivation here is to generate trajectories in the right direction rather than to optimize the speed of movement.
The goal of the unsupervised as well as the error-signal based learning algorithm is to minimize
the summed squared error,
\begin{equation}
\label{serrEq}
\xi_n = \sum_{k=1}^n \lambda^{n-k} e^2_k, 
\end{equation}
where $e_k = (\ell_k - \hat{\ell}_k) =  (\ell_k - {\beta_k^\prime}^\mathrm{T} w_n)$. $n$ is the index of the current time step and $k$ is an index over past time steps. $\lambda$ is a constant 
for degrading the relative contribution of the past time steps ($0 < \lambda \leq 1$).
$\xi_n$ can be further expressed as,
\[
\xi_n = \sum_{k=1}^n \lambda^{n-k} (\ell^2_k - 2 w^\mathrm{T}_n \beta^\prime_k \ell_k + w^\mathrm{T}_n  \beta^\prime_k  {\beta_k^\prime}^\mathrm{T}  w_n ).
\]
Optimum parameters can be found by solving
\[
\nabla_{w_n} \xi_n = \sum_{k=1}^n \lambda^{n-k} ( - 2  \beta^\prime_k \ell_k + 2  \beta^\prime_k  {\beta_k^\prime}^\mathrm{T}  w_n ) = 0.
\]
Defining 
\[
\sum_{k=1}^n \lambda^{n-k}  {\beta_k^\prime} \ell_k = \Theta_n \text{     and     } \sum_{k=1}^n \lambda^{n-k}  {\beta_k^\prime}  {\beta_k^\prime}^\mathrm{T}  = \Psi_n,
\]
solution to $ \nabla_{w_n} \xi_n =0$ can be found as
\begin{eqnarray*}
			 \nabla_{w_n} \xi_n &= 0 & = - 2 ( \Theta_n  - \Psi_n w_n)  \\
			 \Rightarrow \Theta_n  		   & = &	 \Psi_n w_n \\
			   	\Rightarrow \hat{w}_n	   & = &	 \Psi^{-1}_n \Theta_n.
\end{eqnarray*}
Utilizing matrix inversion dilemma, Recursive Least Squares (RLS) \cite{Farhang1999} algorithm proposes a recursive formulation for  $\Psi^{-1}$
\[
\Psi^{-1}_n =P_n = \lambda^{-1} (P_{n-1} -k_n {\beta_n^\prime}^\mathrm{T}P_{n-1}),
\]
where
\[
k_n = \frac{P_{n-1} \beta^\prime_n } { \lambda + {\beta_n^\prime}^\mathrm{T} P_{n-1} \beta^\prime_n}.\\
\]
\vspace{3mm}

Our method aims at simultaneous harvesting of various decoding parameters $B_d$ and, hence, $\beta$ and detecting optimum meta-parameters $w$. These subtasks correspond to exploration and exploitation phases of a reinforcement learning algorithm, respectively. We employ $\epsilon$-greedy exploration policy. In other words, with a predefined probability, $\epsilon$, the algorithm prefers exploring the parameter space, which means a new $\beta$ is chosen randomly. Otherwise, i.e. with a probability of $1-\epsilon$, the algorithm uses the best decoding parameters, i.e. the beta that minimizes the estimated decoder cost (equation~\ref{estimatedDecoderCost}) . Given $\hat{w}$, the current estimate of $w$, the optimal unit normed $\beta$ is computed by finding $\arg \max_{|\beta| =1} {\beta^\prime}^\mathrm{T} \hat{w}$. This is equivalent to maximizing the cosine between ${\beta^\prime}$ and $\hat{w}$ by setting ${\beta^\prime} = \hat{w}$ and normalizing the corresponding $\beta$. A pseudocode for the algorithm is sketched in table~\ref{t:alg}.
 
%----algorithm RLS

\begin{table}
\centering
\begin{minipage}{0.77\linewidth}
\label{t:alg}
\newcommand{\single}{\hspace*{5ex}}
\newcommand{\double}{\hspace*{10ex}}

\textbf{RLS based algorithm for continual \emph{unsupervised} adaptation of the decoding parameters} \\
\\

\textbf{for} time step $n$ at every $T$ time steps do  \\

\single	\# select B \\
\single	\textbf{if} random  $> \epsilon$ \\
	
\double		$\beta_n  \leftarrow  \arg \max_{|\beta| =1} {\beta^\prime}^\mathrm{T} \hat{w}_{n-T}$ \\
 	
\single	\textbf{else} \\
	\\
\double		 $\beta \rightarrow$ random \\

\single	\textbf{endif} \\

\single	\# make prediction on $-log$-cost\\
\single	$\hat{\ell}_n =  {\beta_n^\prime}^\mathrm{T} \hat{w}_{n-T}$ \\ 
\single	 \\
 
\single	\#observe the $-log$-cost of the last $T$ time steps from user's $u_t$ \\
\single	$\ell_n =  -\log(J^d_n) = -\log(\sum_{t=n-T+1}^n  u^\mathrm{T}_t  u_t) $ \\
\double		\#or alternatively according to equation~\ref{errorCost}\\
\single	 \\
\single	\#compute the prediction error \\
\single	$e_n = \ell_n -\hat{\ell}_n$ \\
	
\single	\#compute the innovation gain \\
\single	$k_n = \frac{P_{n-T} \,  \beta^\prime_n } {\lambda +  {\beta_n^\prime}^\mathrm{T} \, P_{n-T} \,  \beta_n^\prime}$ \\
	
\single	\#update meta parameters \\
\single	$\hat{w}_n = \hat{w}_{n-T} + k_n \, e_n $  \cite{Farhang1999}\\

\single	\#update inverse of the correlation matrix \\
\single	$P_n = \lambda^{-1} (P_{n-T} - k_n \, {\beta_n^\prime}^\mathrm{T} \, P_{n-T})$ \\

\textbf{endfor}\\[2mm]
\end{minipage}
\caption{A sketch of the unsupervised learning algorithm via RLS}
\end{table}

\subsubsection{Adaptive supervised recursive least squares filtering}
\label{sec:sup}
%NOW SUPERVISED ALGORITHM
%----algorithm RLS SUPERVISED

\begin{table}
\centering
\begin{minipage}{0.77\linewidth}
\newcommand{\single}{\hspace*{5ex}}
\newcommand{\double}{\hspace*{10ex}}
\textbf{RLS algorithm for \emph{supervised} adaptation of the decoding parameters} \\
\\

\textbf{for} every time step $t$  do  \\

\single	\# make prediction on $v^{intent}_t$\\
\single	$\hat{v}^{intent}_t = B^\prime_d \, u_t$\\ 
 
\single	\#observe user's $v^{intent}_t$ \\

\single	\#compute the prediction error \\
\single	$e^{intent}_t = v^{intent}_t - \hat{v}^{intent}_t $\\
	
\single	\#compute the innovation gain \\
\single	$k_t = \frac{P^u_t \, u_t } {\lambda_{sup} +  u_t^\mathrm{T} \, P^u_t \, u_t}$ \\
	
\single	\#update $B^\prime_d$ (and hence $B_d$) matrix \\
\single	$B^\prime_d \leftarrow B^\prime_d  +  e^{intent}_t \: {k_t}^\mathrm{T}$ \\

\single	\#update inverse of the correlation matrix \\
\single	$P^u_{t+1} = \lambda_{sup}^{-1} (P^u_t - k_t \, u_t^\mathrm{T} \, P^u_t)$ \\

\textbf{endfor}\\[2mm]
\end{minipage}
\label{t:supalg}
\caption{A sketch of the supervised RLS algorithm}
\end{table}

Under the assumption that the decoder knows the intended movements of the user, $B_d$ can be adapted to $B_u$ by utilizing an RLS filter. 
Let $v^{intent}$ be the intended velocity of the user at time step $t$.
\[
v^{intent}_t = B^\prime_u \, u_t,
\]
where $B^\prime_u$ is the submatrix of the third and fourth rows of $B_u$, i.e. $B^\prime_u= \left( \begin{array}{r} B_{u3}\\B_{u4}\end{array} \right)$. The supervised decoder estimates the intended velocity using $B_d$, 
\[
\hat{v}^{intent}_t = B^\prime_d \, u_t.
\]
For the supervised decoder, it is assumed that the user's intent, $v^{intent}$, is known to the decoder. The supervised RLS learning algorithm infers $B_u$ online from $v^{intent}_t - \hat{v}^{intent}_t$.
The supervised adaptive decoder is used to benchmark the proposed unsupervised and error-based adaptive decoders. The supervised RLS method is described in table~\ref{t:supalg}. Note that $P^u_t$ stands for the inverse of the $C \times C$ sample correlation matrix for $(u_0 ... u_t)$. $\lambda_{sup}$ is the forgetting parameter of the supervised algorithm and is set to $1$. $P^u_0$ is set to $100\,I$.

\subsection{Simulation Procedures}
\label{sec:sim}
We simulated the interaction of the optimal feedback controller with different adaptive decoders described in section~\ref{sec:methods}. The behavior of the BMI user was simulated using the framework of stochastic optimal feedback control which has been shown to provide a good model for human motor behavior in various motor tasks \cite{Todorov2002, Diedrichsen2010, Braun2009}. The combined system of  the optimal controller and the adaptive decoder was simulated at $40$~ms time steps and we used a sensory delay of $200$~ms. The user's task was to control a mouse cursor. The user selects a target at $0.2$~m distance with a random orientation at each trial. The user has to reach the target within $4$~seconds and stay at the target for at least $0.16$ seconds. Upon both success or failure, the user selects a new target. The distance cost parameter $q$ and control signal cost parameter $r$ are both set to $0.02$. We set $\Omega_{\rho}$ to $8\times10^6\,I$, so that the cursor speed-noise had an average standard deviation of $0.0625$~m/s over a uniform distribution of unit normed $\beta$ vectors. The variance value was manually adjusted to obtain the aimed speed noise by testing on $10^4$ unit normed random $\beta$ vectors.

Note that the decoder does not have the information whether a trial is finished or continuing, nor does it know the target of the cursor movement. We simulated and evaluated the following adaptive decoders:
\begin{description}
\item [Unsupervised:] The decoder learns exclusively from continuous neuronal control signals of the user according to equation~\ref{usCost}, without any additional information. Note that the decoder knows neither whether the target has been reached nor when a trial finishes.
\item [Error signal based:] The adaptation uses binary neuronal error signals which indicate the time points where the decoded movement deviates from the intended movement more than $20^{\circ}$. The reliability of the neuronal error signal was mainly assumed to be $80\%$, i.e. swapping probability, $\kappa$, was $0.2$. The effect of various $\kappa$ on the decoding performance, however, was also investigated in section \ref{sec:esr}. 
\item [Unsupervised + error signal based:] The combination of the unsupervised and the error-signal based decoders, i.e. $\ell_n$ was a linear combination of the unsupervised $\ell_n$ and the error-signal based $\ell_n$.
\end{description}

For all of the above algorithms, the current cost is computed from the last $100$ time steps ($T=100$). This corresponds to a parameter update period of $4$~seconds. $\lambda$ of equation~\ref{serrEq} was set to $1$, i.e. no gradual discount of the parameter history was performed. Exploration rate was  $0.4$, i.e. $\epsilon= 0.4 $. We simulated $50$ random instantiations of all these unsupervised and the error signal based adaptive decoding algorithms. 1501 successive trials of target reaching were simulated for each instantiation. Note that from trial 1463 on the adaptation of the decoding algorithms was frozen and the current optimal decoding parameters were used for the last 39 trials (decoder-freeze). We evaluated their performances and compared it to the performance of a supervised adaptive decoder where the adaptation is based on perfect knowledge of the intended movement velocity at each time step (see section~\ref{sec:sup}). Such a supervised adaptive decoder yields the best possible adaptation, however, it assumes knowledge that is certainly not available during autonomous BMI operation. In addition, we also compared the performance of our adaptive decoders to the performance of a static untrained random decoder.

\begin{figure}
  \centering
    \includegraphics[width=0.73\textwidth]{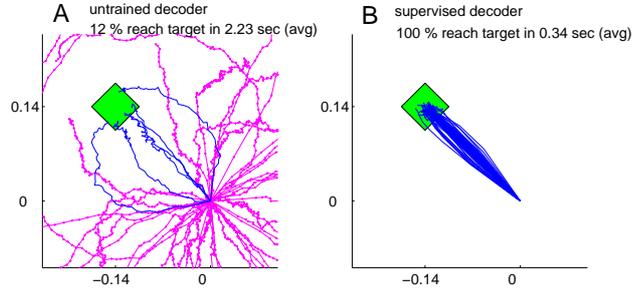}    
  \caption{The trajectories for random decoders (A) and supervised adaptive decoders (B) during decoder-freeze, i.e. after decoder exploration and adaptation have been switched off for performance evaluation. Magenta thick curves indicate the failed trajectories. Each plot depicts the trajectories of 50 training simulations, each at trial 1501. The 50 different targets and trajectories at trial 1501 are rotated to the same orientation ($\frac{3\pi}{4}$) for a better visual evaluation.}
 \label{trjBasl} 
\end{figure}	

\begin{figure}
  \centering
    \includegraphics[width=0.99\textwidth]{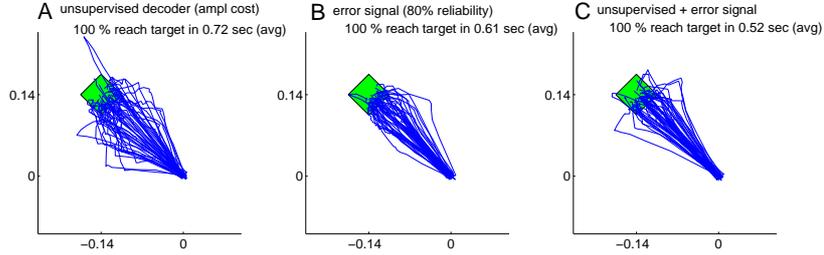}    
  \caption{The trajectories for different strategies and their variations during decoder-freeze. Magenta thick curves indicate the failed trajectories. Each plot depicts the trajectories of 50 training simulations, each at trial 1501. Note that not only trial 1501 included 50 simulations, but the whole history of 1501 trials are simulated 50 times with random initial tunings. The 50 different targets and trajectories at trial 1501 are rotated to the same orientation ($\frac{3\pi}{4}$) for a better visual evaluation.}
 \label{trjGen} 
\end{figure}

\section{Results}
\label{sec:res}
Our findings show that all the decoders described in section~\ref{sec:sim} can rapidly adapt to accurate cursor control from totally unknown tuning of the neuronal signals to movement velocity whereas the random decoder fails to reach the target (figure~\ref{trjBasl}A). Although trajectories of the unsupervised and error-based decoders after adaptation are more jerky compared to the supervised case, they are still mainly straight and yield a high target hit rate of nearly 100\% (figure~\ref{trjGen}). These results show that decoders can be trained during autonomous BMI control in the absence of any explicit supervision signal.
 
\begin{figure}
  \centering
    \includegraphics[width=0.99\textwidth]{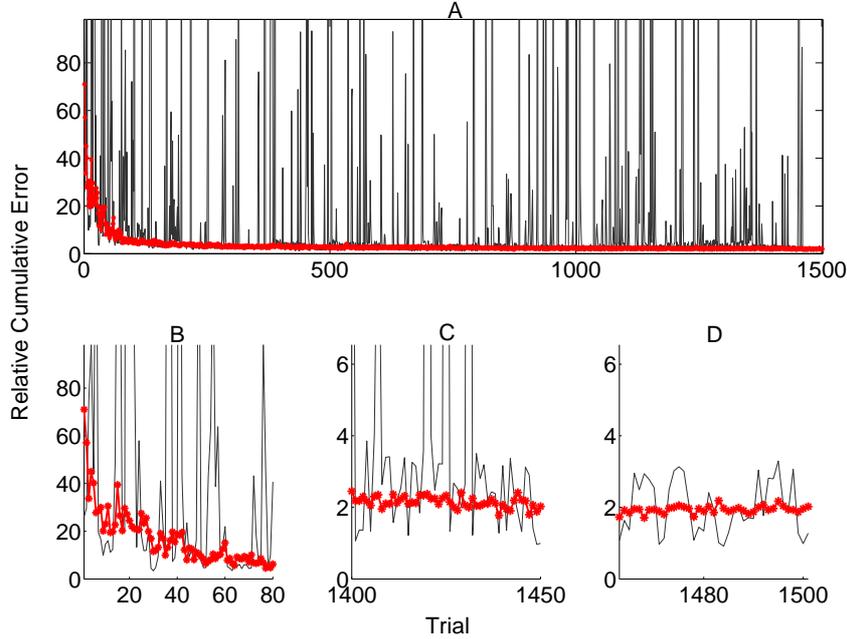}     
  \caption{Evolution of the relative cumulative error (RCE)  for the unsupervised strategy (A). The plot shows the RCE with respect to trial number for a single simulation (gray) and the Median RCE (MRCE) for 50 simulations with random initial tuning parameters (red with marker). Zoom into RCE and MCE curves for early (A) and late (B) learning and during decoder-freeze (C).}
 \label{bgSamples} 
\end{figure}

\begin{figure}
  \centering
    \includegraphics[width=0.99\textwidth]{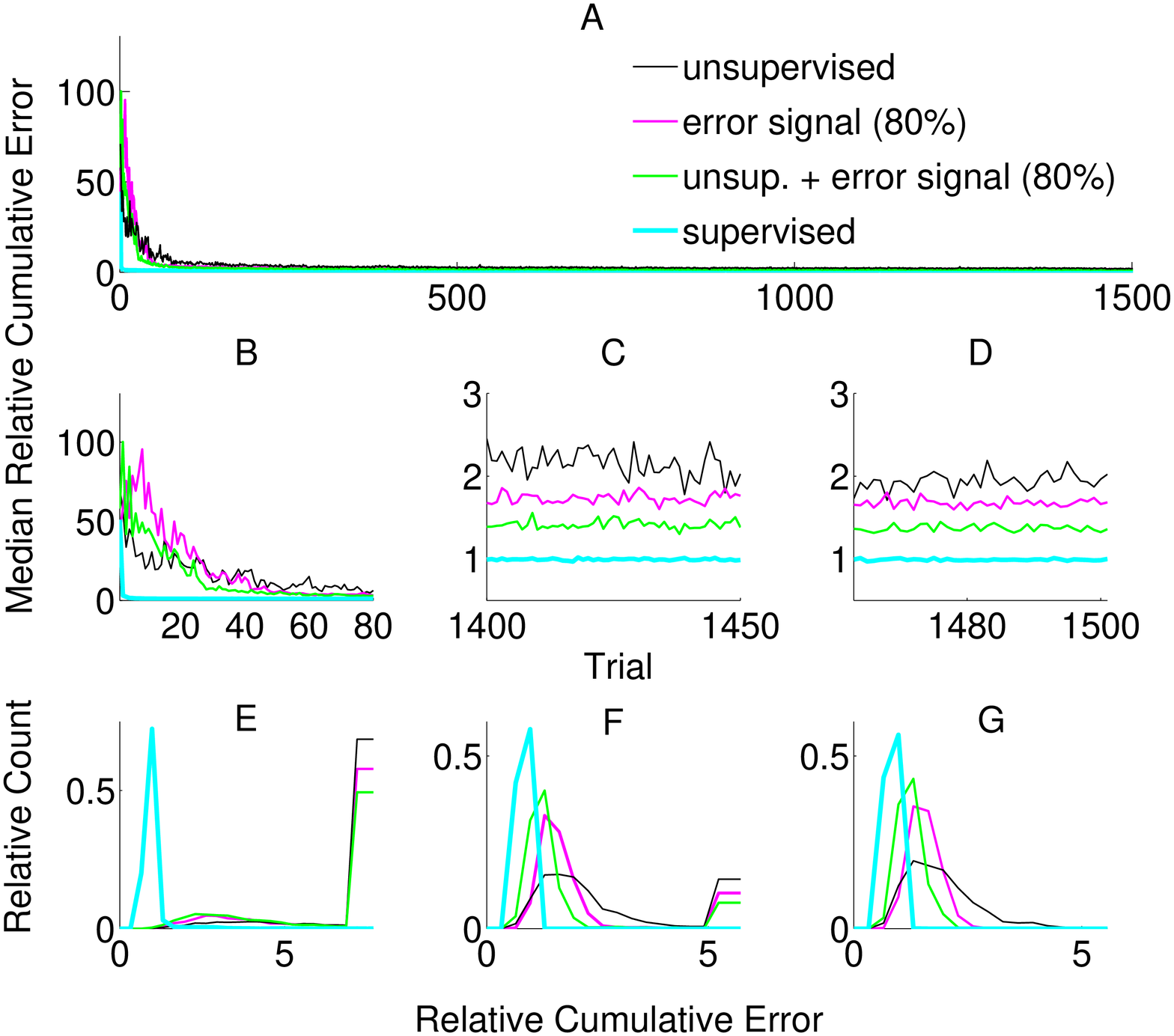}  
  \caption{Comparison of the unsupervised (black), the error signal based ($80\%$ reliability, magenta), their combination (green) and the supervised (cyan) strategies. Medians of relative cumulative errors of 50 simulations from each group for all of the trials (A). Zoomed medians for early (B) and late (C) learning and during decoder-freeze (D). The distributions of the relative cumulative errors for each of the phases (E, F, G). The rightmost values of the distribution plots denote the total relative counts of the outlier values that are greater than the associated $x$-axis value. The number of outlier values decreased across trials, i.e. it was the highest during early learning and zero during decoder-freeze. Outliers correspond to the failed trajectories in figures~\ref{trjGen} and ~\ref{trjGENUFR}.}
 \label{USerror} 
\end{figure}

\subsection{Comparison of different adaptation algorithms}
As a baseline for comparisons, we implemented the supervised decoder that knows the intention of the user and fits the decoder parameters, $B_d$, based on this intention (see section~\ref{sec:sup}). Though unrealistic, this learning scheme is obviously the most successful of the presented methods (figure~\ref{trjBasl}B). In order to compare different algorithms, we utilize a measure that counts for the cumulative distance to the movement target and call it \emph{cumulative error},
\[
\text{Cumulative Error }(m) = \sum_{t=1}^{M_m}  \|g_t -p_t\|,
\] 
where $g_t$ and $p_t$ and are the $2$-dimensional target and cursor position vectors at time step $t$ of trial $m$, respectively. $M_m$ is duration of trial $m$ in time steps. For a more intuitive interpretation of the given error measure, we present out results in terms of \emph{relative cumulative error}, which is the normalized cumulative error with respect to average cumulative error of the supervised decoder after adaptation,
\[
\text{Relative Cumulative Error }=  \frac {\text{Cumulative Error}}{\text{Mean Supervised Cumulative Error}}.
\]

\begin{figure}
  \centering
    \includegraphics[width=0.73\textwidth]{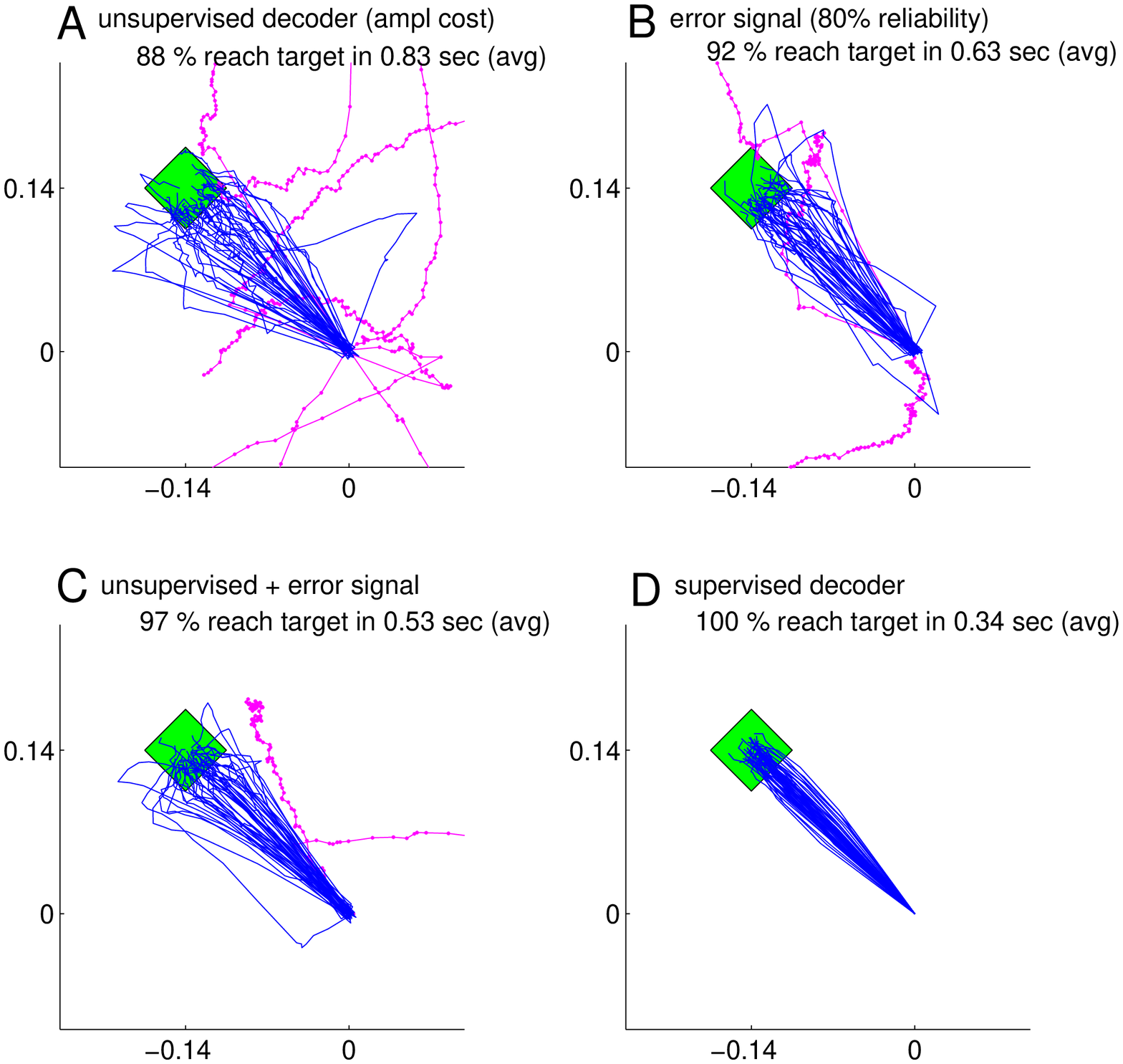}    
  \caption{The trajectories for different strategies and their variations in the late learning phase. Magenta thick curves indicate the failed trajectories. Each plot depicts the trajectories of 50 training simulations, each at trial 1404. Durations and target hit rates, however, are computed from pooled trajectories of 5 consecutive trials (1402-1406, totally 250 trajectories).
}
 \label{trjGENUFR} 
\end{figure}

Figure~\ref{bgSamples}A depicts the evolution of the relative cumulative error for a single simulation (gray) of the \emph{unsupervised} algorithm and median relative cumulative errors (MRCE) for $50$ randomly initialized simulations (red, *). The jumps in the gray curve correspond to exploration periods, where random decoder matrices, $B_d$, are explored and evaluated. The relative cumulative error shows different characteristics for early learning (\ref{bgSamples}B), late learning(\ref{bgSamples}C) and decoder-freeze (\ref{bgSamples}D) phases. The early learning phase was investigated to compare the learning speeds of different algorithms, whereas the late learning phase shows the \emph{saturated} final performance of the algorithms when adaptation continues. During decoder-freeze (last 39 time steps), the adaptation (also the exploration) was stopped and the final performance of the decoder was evaluated. No jumps in the performance are observed anymore due to absence of exploration.

Figure~\ref{USerror} shows the comparison between supervised, error signal based and unsupervised algorithms. MRCEs across 50 runs are shown for the entire simulation (A), for early learning (B), for late learning (C) and during decoder-freeze (D) (figure~\ref{USerror}). Distribution of the relative cumulative errors for the individual phases (figure~\ref{USerror} E, F, G) reveal that the supervised algorithm is superior to the other algorithms in all phases ($p<0.01$, Wilcoxon rank sum test). In all of the phases,  the combination of the error signal based and the unsupervised strategies yielded a significantly lower cumulative error than the individual strategies alone ($p<0.01$, Wilcoxon rank sum test). The performance of the the error-based learning was significantly better than the unsupervised strategy also for all of the phases ($p<0.01$, Wilcoxon rank sum test). Note that the trajectories reached the targets not only during decoder-freeze (figure~\ref{trjGen}) but also mostly in the late learning phase (figure~\ref{trjGENUFR}), where exploration can occasionally cause some trajectories to deviate strongly from a straight line towards the target.

\begin{figure}
  \centering
    \includegraphics[width=0.99\textwidth]{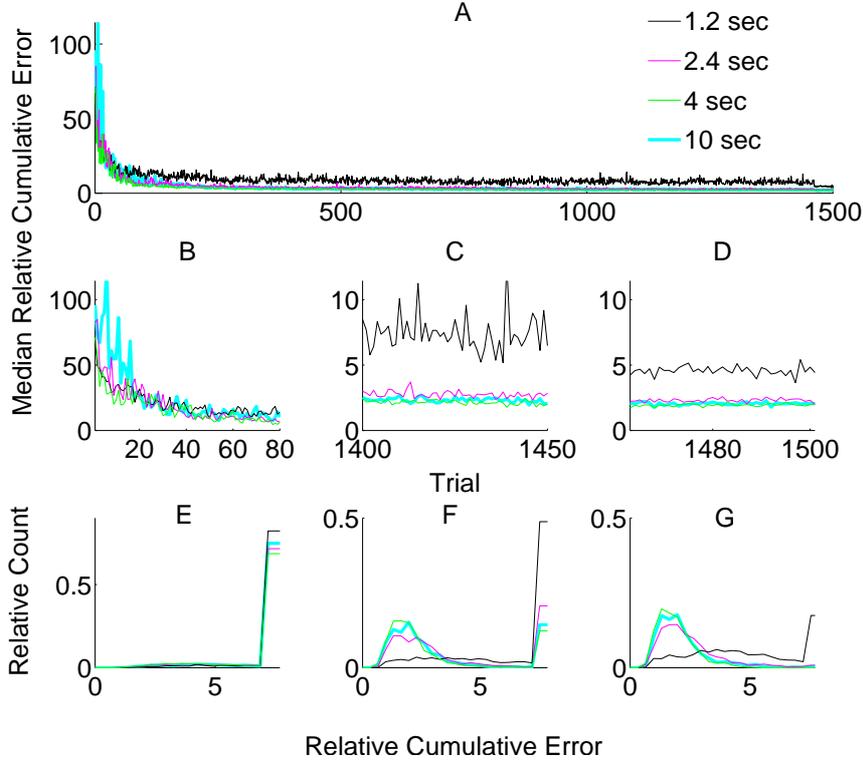}    
  \caption{Comparison of relative cumulative error measures using the unsupervised strategy for different update periods ($T$). The algorithm updates the decoding parameters either every $1.2$ seconds (black) or  $2.4$ seconds (magenta) or  $4$ seconds (green) or  $10$ seconds (cyan). Medians of relative cumulative errors of 50 simulations from each group for all of the trials (A). Zoomed medians for early (B) and late (C) learning and during decoder-freeze (D). The distributions of the relative cumulative errors for each of the phases (E, F, G). The rightmost values of the distribution plots denote the total relative counts of the outlier values that are greater than the associated $x$-axis value. The number of outlier values decreased across trials, i.e. it was the highest during early learning and the lowest during decoder-freeze.}
 \label{allts} 
\end{figure}

\subsection{Effect of the parameter update period}
We varied the parameter update period, $T$, between $1.2$ and $10$ seconds in order to check the stability of the unsupervised strategy with respect to this parameter (figure~\ref{allts}). Our results show that for all tested update periods greater than or equal to $2.4$ seconds, the performance depended only weakly on exact value of the update period. Though the performance for an update period of $2.4$~s was significantly ($p<0.05$, Wilcoxon rank sum test) inferior compared to an update rate of $4$~s or $10$~s in both late learning and during the freeze of the decoder, the difference was minor. Moreover, the performance of the update periods $4$~s and $10$~s were not significantly different during decoder-freeze ($p>0.05$, Wilcoxon rank sum test). We, therefore, conclude that our algorithm is robust against the update rate as long as it is high enough and used an update period of $4$~seconds for all remaining simulations.

\begin{figure}
  \centering
    \includegraphics[width=0.99\textwidth]{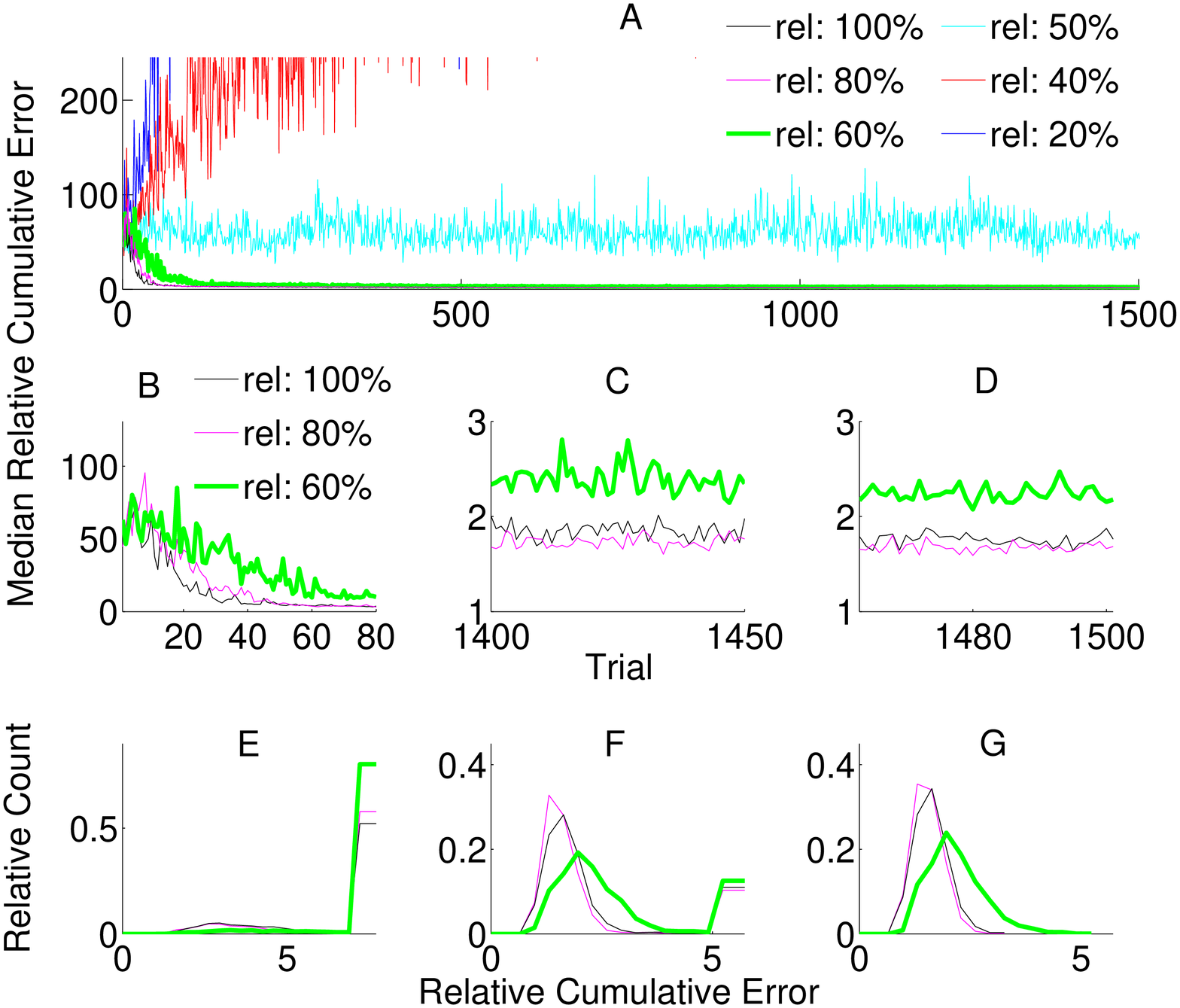}    
  \caption{Comparison of relative cumulative error measures using the error-signal-based training strategy for different reliabilities of error signals. 
 The error signal reliability is either $100\%$ (black) or  $80\%$  (magenta) or  $60\%$  (green) or  $50\%$  (cyan) or  $40\%$  (red) or  $20\%$  (blue). Medians of relative cumulative errors of 50 simulations from each group for all of the trials (A). Zoomed medians for early (B) and late (C) learning and during decoder-freeze (D). The distributions of the relative cumulative errors for each of the phases (E, F, G). The rightmost values of the distribution plots denote the total relative counts of the outlier values that are greater than the associated $x$-axis value. The number of outlier values decreased across trials , i.e. it was the highest during early learning and the lowest during decoder-freeze. Outliers correspond to the failed trajectories in figures~\ref{trjGen} and ~\ref{trjGENUFR}.
In general, higher the reliability, better the performance. A minimum reliability of $60\%$ is needed for successful training. $100\%$  and  $80\%$ reliabilities are statistically equivalent during decoder-freeze (rank sum test, $p > 0.05$).}
 \label{allerror} 
\end{figure}

\subsection{Effect of Error Signal Reliability}
\label{sec:esr}
Error signal based decoder performance obviously depends on the reliability of the error signals. Our results so far used an error signal with $80\%$ reliability, i.e. $\kappa = 0.2$. Although several studies have shown evidence on neuronal error signals \cite{Falkenstein2000, Gehring1993, Diedrichsen2005, Krigolson2008}, conclusive quantitative data on the reliability of the error signals is still missing. To compute the dependence of the error-based adaptive decoder on $\kappa$, we varied it between $0$ and $0.8$. Our findings show that the reliability must be greater than $50\%$ for successful adaptation (figure~\ref{allerror}A,B,C,D). Reliabilities of $80\%$ and $100\%$ yielded statistically indistinguishable performance during decoder-freeze ($p>0.05$, Wilcoxon rank sum test). Though $80\%$ was slightly yet significantly better than $100\%$ in the late learning phase ($p=0.049$). A decoder with a reliability of $60\%$ yielded a significantly inferior performance in all of the phases to the decoders with $80\%$ and $100\%$ reliability (figure~\ref{allerror}E, F, G, rank sum test, p-value $< 0.05$), its median error during late learning and freezing was only about $30\%$ higher. Decreasing the reliability further to $50\%$ drastically increased the median relative cumulative error.

\begin{figure}
  \centering
    \includegraphics[width=0.99\textwidth]{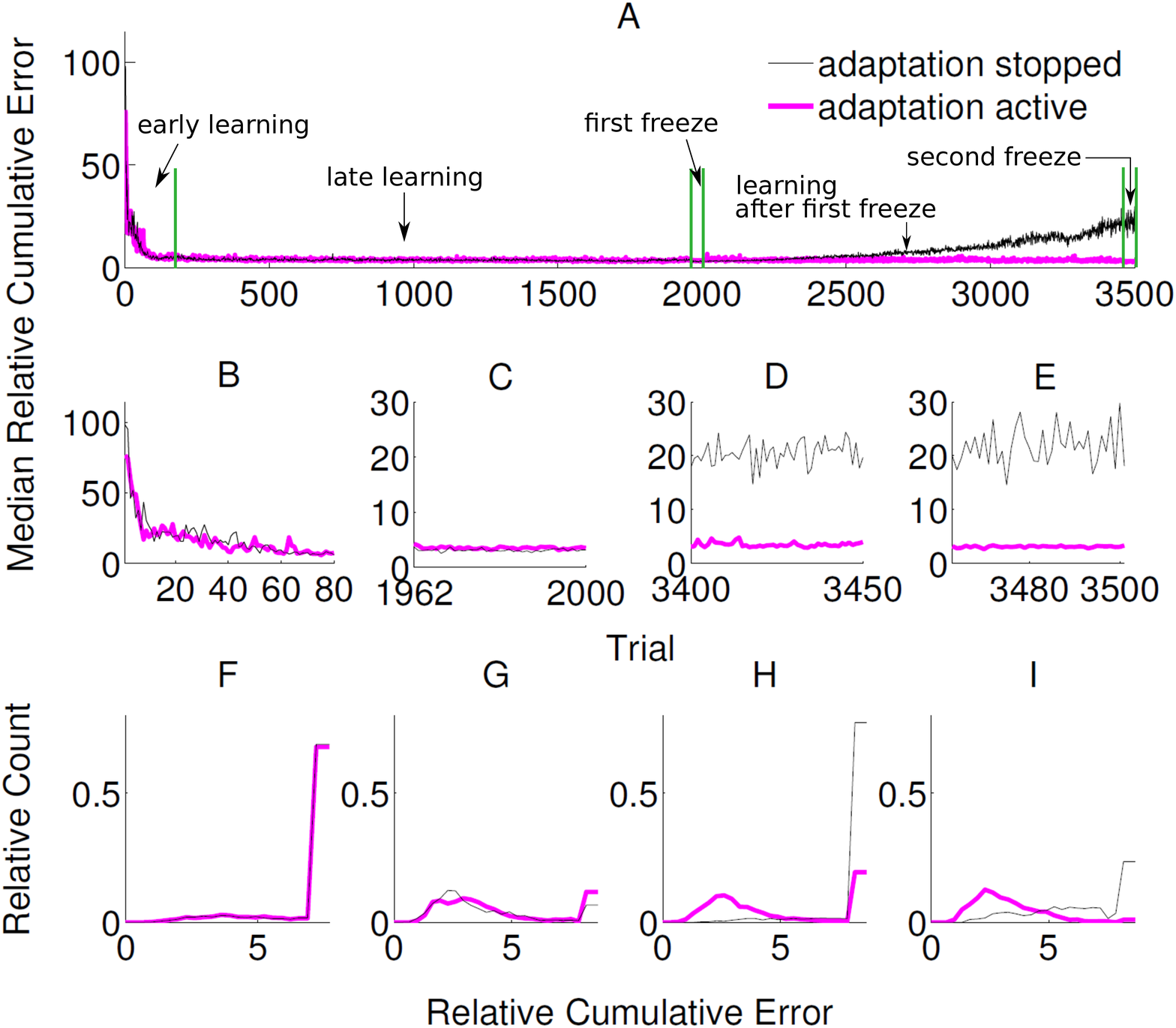}    
  \caption{Relative cumulative errors for the unsupervised strategy under nonstationary tuning. Black curve shows the median RCE of 50 simulations, where adaptation was active between trial 1 and 1961 and stopped after that. Magenta curve shows the median RCE of 50 simulations, where adaptation was active \emph{both} between trial 1 and 1961 and after trial 2000. In both groups, adaptation was inactive between 1962 and 2000 for comparison purposes (A). Zoomed medians during the early learning phase (B), the first decoder-freeze (C), the late learning phase after the first freeze (D) and the second decoder-freeze (E). The distributions of the relative cumulative errors for each of the phases (F, G, H, I). The rightmost values of the distribution plots denote the total relative counts of the outlier values that are greater than the associated $x$-axis value. Outliers correspond to the failed trajectories in figure~\ref{trjRW}.}
 \label{rwfig} 
\end{figure}

\begin{figure}
  \centering
    \includegraphics[width=0.99\textwidth]{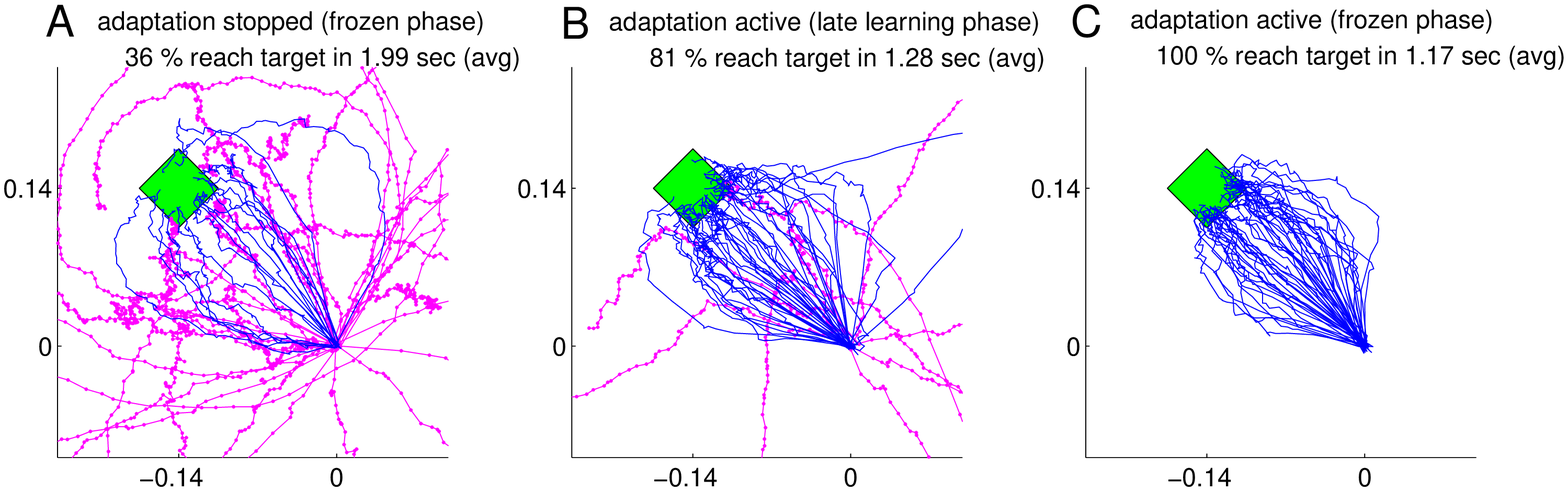}    
  \caption{The trajectories under nonstationary tuning for unsupervised strategy and their variations during late learning and during decoder-freeze. Magenta thick curves indicate the failed trajectories. Each plot depicts the trajectories of 50 training simulations. Trajectories during decoder-freeze belong to trial 3501. The late learning trajectories were recorded at trial $3402$, durations and target hit rates are from pooled trajectories of trials $3402$-$3406$. The 50 different targets and trajectories at the recorded trial are rotated to the same orientation ($\frac{3\pi}{4}$) for a better visual evaluation.}
 \label{trjRW} 
\end{figure}	

\subsection{Adaptivity to nonstationary tuning}
We furthermore investigated, whether the unsupervised adaptive algorithm can cope with continual changes in the tuning. The velocity tuning parameters of the user, i.e. $\beta_u$, flattened third and fourth rows of $B_u$, were changed after each trial according to the following random walk model,
\[
\beta_u \leftarrow \beta_u + \varrho,
\]
where $\varrho$ is $40$-dimensional row vector whose entries are randomly drawn from a normal distribution, $\varrho \sim \mathcal{N}(0,0.007)$.  
We put a hard limit on the magnitude of the entries of $\beta_u$, so that they did not exceed $-0.3$ and $0.3$. In order to investigate the performance of our algorithm under nonstationary tuning, 50 randomly initialized \emph{unsupervised} adaptive decoders was  compared to another group of 50 randomly initialized \emph{unsupervised} decoders, for which adaptation was stopped after a certain amount of trials.
Both decoder groups were adaptive for the first $1961$ trials, at the end of which they reached a stationary performance (figure~\ref{rwfig}A,B,F). Then, the adaptation of both groups was switched off during trials $1962$ to $2000$ (1st freeze) to compare the baseline performance of both decoder groups (figure~\ref{rwfig}C,G). As expected, both groups performed equally well during the first $2000$ trials ($p>0.1$, Wilcoxon rank sum test). For the first group, the adaptation was then switched on again for the next $1462$ trials, whereas for the other group the adaptation was kept off. Evidently, non-adaptive decoders could not cope with the changing tuning anymore and the performance strongly degraded (figures~\ref{rwfig}A,D,H and \ref{trjRW}A). Adaptive decoders, in opposite, tracked the changes in $B_u$ well and kept the performance stable (figure~\ref{rwfig}A,D,H). After trial $3462$ a 2nd freeze phase of $39$ trials was used to quantify the difference in performance between both groups for nonstationary tuning parameters: adaptive decoders yielded a significantly ($p<0.0001$ Wilcoxon rank sum test) and about $7$ times smaller error than non-adaptive decoders (figure~\ref{rwfig}A,E,I). In these simulations, we employed the unsupervised decoder cost as in equation~\ref{usCost}. The simulation settings are the same as described at the beginning of section~\ref{sec:res} except for $\lambda$. Here, we set $\lambda = 0.995$, to reduce the influence of the earlier trials relative to the recent ones. This improves performance as recent trials contain relatively more information on $B_u$. Trajectories of the adaptive group reach very accurately to the target during decoder-freeze (figure~\ref{trjRW}C) and less but also with high accuracy during the late learning phase (figure~\ref{trjRW}B).

\begin{figure}
  \centering
    \includegraphics[width=0.99\textwidth]{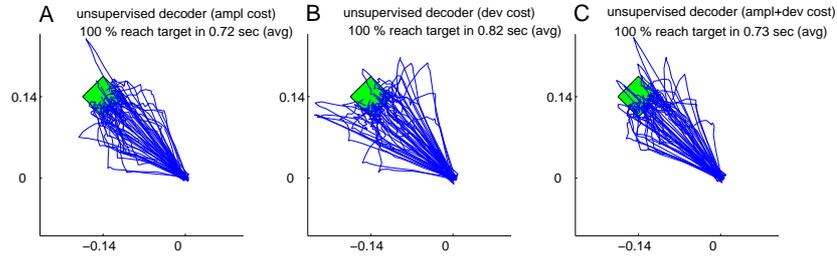}    
  \caption{The trajectories for different cost measures and their variations during decoder-freeze. Magenta thick curves indicate the failed trajectories. Each plot depicts the trajectories of 50 training simulations, each at trial 1501. Note that not only trial 1501 included 50 simulations, but the whole history of 1501 trials are simulated 50 times with random initial tunings. The 50 different targets and trajectories at trial 1501 are rotated to the same orientation ($\frac{3\pi}{4}$) for a better visual evaluation.}
 \label{trjCost} 
\end{figure}	

\begin{figure}
  \centering
    \includegraphics[width=0.99\textwidth]{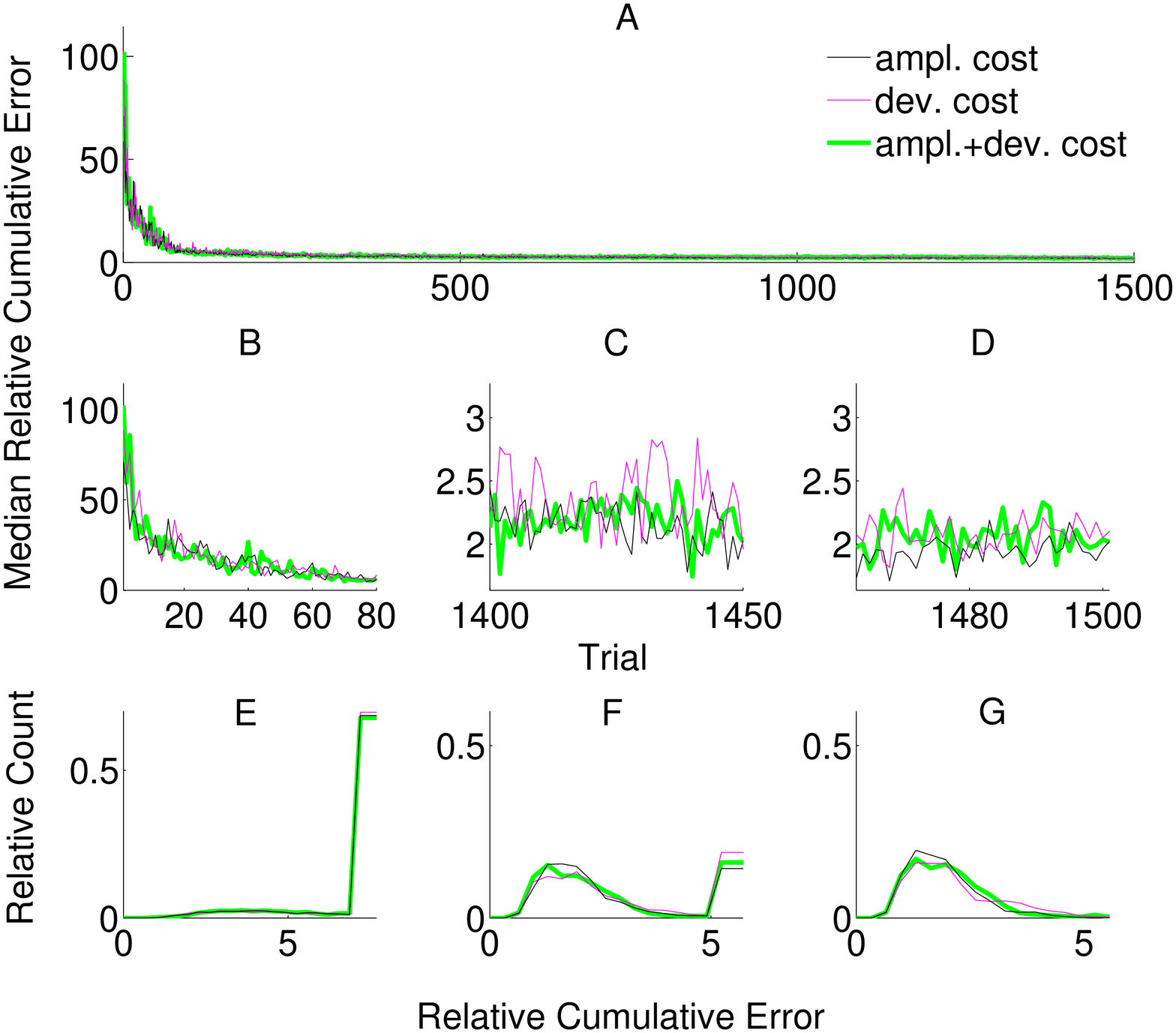}    
  \caption{Comparison of relative cumulative errors for the unsupervised strategy using different cost measures. The algorithm employs either amplitude cost (black), or deviation cost (magenta) or their combination (green). Medians of relative cumulative errors of 50 simulations from each group for all of the trials (A). Zoomed medians for early (B) and late (C) learning and during decoder-freeze (D). The distributions of the relative cumulative errors for each of the phases (E, F, G). The rightmost values of the distribution plots denote the total relative counts of the outlier values that are greater than the associated $x$-axis value. The number of outlier values decreased across trials, i.e. it was the highest during early learning and zero during decoder-freeze.}
 \label{allcost} 
\end{figure}

\subsection{Different decoder costs for unsupervised adaptation}
Figure~\ref{trjCost}B shows the trajectories obtained by 50 simulations of the unsupervised algorithm using deviation cost (equation~\ref{usCost2}) during decoder-freeze. The trajectories were precise and fast. 

The trajectories obtained using amplitude + deviation cost (equation~\ref{usCost3}) are shown in figure~\ref{trjCost}C. Again, straight and fast movements were obtained. A comparison between the three unsupervised cost functions is presented in figure~\ref{allcost}. All these three costs yielded equal performance ($p>0.05$ Wilcoxon rank sum test) during all phases.

\section{Discussion}
Our results show that under realistic conditions, adaptive BMI decoding starting with random tuning parameters is feasible without an explicit supervision signal. Decoding performance gradually improves across trials and reaches a value close to the maximum possible performance as obtained by a supervised adaptive decoder, which assumes perfect knowledge of the intended movement of the BMI user. Moreover, we propose an adaptive decoder which employs neuronal error signals and show that this decoder yields a similar performance to our unsupervised adaptive decoder. Unsupervised and error-signal based decoders adapt rapidly and generate precise movement trajectories to the target. The suggested decoders do not require a supervision signal, e.g. the intended movement, and therefore can be used during autonomous BMI control. The suggested unsupervised adaptation is based on the minimization of a simple cost function, which penalizes high control signals and/or high variability of the neuronal control signals. The rationale behind these costs is, that inaccurate decoding causes corrective attempts by the BMI user, which in turn increase control signals and control signal variability. Therefore, accurate movement decoding corresponds to lower costs and the minimization of the suggested cost functions improves the accuracy of BMI movement control. Due to the generality of this approach we expect this to work in different kinds of motor tasks and not only for the reaching task considered in our simulations. Note that the cost function could be alternatively derived from only trajectories 
instead of control signals (e.g. deviations from straight line could be punished). An additional argument in favor of our cost functions comes from behavioral studies of human motor control: A wide range of human motor behavior can be described by optimal feedback control (OFC) models, which minimize cost functions containing the same dependence on the motor control signals as we used in our decoder cost \cite{Todorov2002, Braun2009, Diedrichsen2010}.

Besides adaptation to unknown but static neuronal tuning to movement, we demonstrated that the proposed algorithms can also keep the decoding performance stable for nonstationary tuning. This is even possible if the tuning is not only nonstationary but also initially unknown. In these simulations, we assumed that the nonstationarity of the tuning parameters follows a random walk model and, hence, is independent of the decoder. If the decoded movement is fed back to the BMI user, the neuronal signals might adapt \cite{Jarosiewicz2008} and the learning speed as well as the final accuracy might even increase beyond the presented values.

In order to train the decoder, we assumed a log-linear model that relates the decoder parameters to cost via meta-parameters. We introduced a learning algorithm, which explores the parameter space with a $\epsilon$-greedy policy. Our method performs least squares regression recursively to estimate the optimal values of the meta-parameters. In other words, the algorithm performs simultaneous exploration of the decoding parameters and recursive least squares (RLS) \cite{Farhang1999} regression on the decoder cost function. The same algorithm works also with neuronal error signals, where the cost is the number of error signals detected in a given time period. Error related neuronal activity has indeed been recorded from the brain via EEG \cite{Falkenstein2000, Gehring1993, Krigolson2008}, functional magnetic resonance imaging (fMRI)\cite{Diedrichsen2005} and single-unit electrophysiology \cite{Ito2003, Matsumoto2007}. Here, we assume a simple partially reliable error signal that indicates a substantial deviation from the movement intention. Neuronal activity related to this kind of movement execution errors has been found in ECoG \cite{Milekovic2012} and in fMRI \cite{Diedrichsen2005}. Milekovic et al. \cite{Milekovic2012} observed neuronal responses evoked by a $180^{\circ}$ degree movement mismatch during continuous joystick movement in 1-dimension. In our simulations of $2$-dimensional movements, we assumed that neuronal error signals are evoked when the deviation between intended and decoded movement exceeds the somewhat arbitrary threshold of $20^{\circ}$. Although it remains to be shown in future studies that neuronal error signals are indeed observable already at this threshold, we consider this a plausible assumption and expect our algorithm to be robust against the exact value of the threshold. Our results show that the overall performance of our algorithm is robust against different parameter update periods ($T$) and different error signal reliabilities ($>50\%$). Arguably, the proposed algorithm has the potential to work with various types of neuronal error signals, though the computation of the cost function in terms of error signals might need adjustments to achieve high performance.

An alternative to our algorithm would be to use standard reinforcement learning algorithms and generalization methods \cite{Sutton1998} for directly training the decoding parameters without using a meta-parametric model relating cost to decoding parameters. In our practical experience, keeping a record of the previously explored parameters via $P$ matrix of the RLS algorithm and relating the parameters to the $\log$-cost yields good performance. A comparison of our method to different reinforcement algorithms that utilize the same cost and/or other cost functions than the ones suggested here, are interesting topics for future studies. Previously, Kalman filtering methods were also applied for unsupervised adaptation during trajectory decoding~\cite{Eden2004,Eden2004a,Wang2008,Wu2008}. These methods adapt by maintaining consistency between a model of movement kinematics and a neuronal encoding model. They have been shown to track nonstationarities once an initial model is learned via supervised calibration~\cite{Eden2004,Eden2004a,Wang2008}. Our unsupervised approach in this work is fundamentally different from these methods. We assume that, in the aftermath to decoding errors, the statistics of the control signals change and this change is utilized for unsupervised adaptation. In the future, it would be worthwhile to compare the performance of these different methods and their robustness against model violations in online BMI taks.

In principle, our adaptive decoding framework is independent of the type of neuronal signal that is used to control the movement. As neuronal control commands, the instantaneous firing rates of multiple single-unit or multi-unit activities could be used. Alternatively, filtered LFP, ECoG, EEG or MEG signals or the power of LFP, ECoG, EEG and MEG signals in different frequency bands could be employed. Our algorithms assume that neuronal control signals are linearly related to movement velocity. For many different neuronal signal types, indeed, movement trajectories can be reconstructed well using this assumption (\cite{Wessberg2000, Serruya2002, Taylor2002} for SUA, \cite{Pistohl2008, Schalk2007} for ECoG). Linear tuning to movement position or simultaneous linear tuning to position and velocity can easily be implemented in our algorithms by straightforward modifications of the $B$ matrices (see section~\ref{sec:methods}). Future extension of our algorithmic framework might also consider nonlinear tuning. The cost measures we introduced, might need some modifications depending on the tuning of the recorded signals. For instance, if the control signal, e.g. firing rates for individual recording channels, takes an all-or-none behavior, i.e. certain channels are \emph{on} for one direction and \emph{off} for another direction, the norms of the command vectors might hardly vary across different movement directions. In such a case, deviation cost might be preferable over amplitude cost.

\section*{Acknowledgments}
Authors would like to thank German Bundesministerium f\"ur Bildung for grant 01GQ0830 to BFNT Freiburg-T\"ubingen and Boehringer Ingelheim Funds for supporting this work.

%\section*{References}
\bibliography{UnsupervisedAdaptation_GuerelMehring}{}
\bibliographystyle{unsrt}

\end{document}